\documentclass{ieeeaccess}
\usepackage{cite}
\usepackage{amsmath,amssymb,amsfonts}
\usepackage{algorithmic}
\usepackage{graphicx}
\usepackage{textcomp}

\usepackage{amssymb}

\def\BibTeX{{\rm B\kern-.05em{\sc i\kern-.025em b}\kern-.08em
    T\kern-.1667em\lower.7ex\hbox{E}\kern-.125emX}}
\begin{document}
\history{Date of publication xxxx 00, 0000, date of current version xxxx 00, 0000.}
\doi{10.1109/ACCESS.2023.0322000}

\title{Co-attention Graph Pooling for Efficient Pairwise Graph Interaction Learning}
\author{
\uppercase{Junhyun Lee}\authorrefmark{1,*},
\uppercase{Bumsoo Kim}\authorrefmark{2,*},
\uppercase{Minji Jeon}\authorrefmark{3},
and 
\uppercase{Jaewoo Kang}\authorrefmark{1}
}

\address[*]{Equal contribution}
\address[1]{Department of Computer Science and Engineering, Korea University, Seoul 02841, Republic of Korea}
\address[2]{LG AI Research, Seoul 07796, Republic of Korea}
\address[3]{Department of Medicine, Korea University College of Medicine, Seoul 02708, Republic of Korea}
\tfootnote{The research conducted by Jaewoo Kang and Junhyun Lee was supported by grants from the National Research Foundation of Korea (NRF-2023R1A2C3004176), by the Ministry of Science and ICT of Korea (IITP-2023-2020-0-01819) supervised by the Institute of Information \& Communications Technology Planning \& Evaluation, and by the Korea Health Industry Development Institute (HR20C0021(3)).
The research of Minji Jeon was supported by grants from the National Research Foundation of Korea (NRF-2022R1F1A1070111) and the Ministry of Science and ICT of Korea (IITP-2022-RS-2022-00156439) supervised by the Institute of Information \& Communications Technology Planning \& Evaluation.
}

\markboth
{Lee \headeretal: Co-attention Graph Pooling for Efficient Pairwise Graph Interaction Learning}
{Lee \headeretal: Co-attention Graph Pooling for Efficient Pairwise Graph Interaction Learning}

\corresp{Corresponding authors: Jaewoo Kang (kangj@korea.ac.kr) and Minji Jeon (mjjeon@korea.ac.kr).}

\begin{abstract}
Graph Neural Networks (GNNs) have proven to be effective in processing and learning from graph-structured data.
However, previous works mainly focused on understanding single graph inputs while many real-world applications require pair-wise analysis for graph-structured data (e.g., scene graph matching, code searching, and drug-drug interaction prediction).
To this end, recent works have shifted their focus to learning the interaction between pairs of graphs.
Despite their improved performance, these works were still limited in that the interactions were considered at the node-level, resulting in high computational costs and suboptimal performance.
To address this issue, we propose a novel and efficient graph-level approach for extracting interaction representations using co-attention in graph pooling. 
Our method, Co-Attention Graph Pooling (CAGPool), exhibits competitive performance relative to existing methods in both classification and regression tasks using real-world datasets, while maintaining lower computational complexity.

\end{abstract}


\titlepgskip=-21pt

\maketitle

\section{Introduction}
\label{sec:introduction}
\PARstart{R}{ecent} advancements in both aggregation ~\cite{kipf2017semi,velickovic2017graph,gcnii_icml20,pna_neurips20,magna_ijcai21} and pooling operations~\cite{ying2018hierarchical,pmlr-v97-lee19c,Yuan2020StructPool,baek2021accurate,muchpool_ijcai21,sep_icml22,https://doi.org/10.48550/arxiv.2209.02939} have significantly improved the capabilities of Graph Neural Networks (GNNs), enabling for more robust learning of complex graph representations and enhancing performance in downstream tasks like graph classification, node classification, and link prediction.
However, the scope of these approaches are limited on a single graph input while many real-world tasks (e.g., scene graph matching, code search, and drug-drug interaction prediction) require pair-wise analysis of graph-structures.
Therefore, recent studies in GNNs have shifted their focus to representation learning over pairs of input graphs.

One of the earliest approaches for paired graph representation learning using GNNs is the graph convolutional Siamese network~\cite{ktena2017distance}.
In Graph Convolutional Siamese network, the input pairs must share identical graph topology, and the paired training is done by simply concatenating the individual graph representations.
However, since it does not take the interaction between the graphs during the embedding process into account, each graph is embedded into a single \textit{static} representation regardless of its pair.
This static representation can limit the expressiveness of the many-to-many relationships between pairwise graphs~\cite{deac2020empowering}.
For instance, when predicting interactions between chemical compounds, each molecular graph can have multiple functional groups, which are important sub-graphs for the task. 
Since the contribution of each functional group to the interaction depends on its pair, representing molecules with a single static representation may be limited in terms of expressiveness. 
To overcome this limitation, it is necessary to consider the interaction between the input pair of graphs.

Subsequent works~\cite{li2019graph,bai2019simgnn,deac2020empowering} proposed architectures considering the interaction between the input pair using the co-attention mechanism, which is an intuitive way to contemplate pairwise interactions.
While co-attention has improved the predictive power of these methods, they are still limited to obtaining the interaction representation at the \textit{node-level}. 
This not only leads to increased complexity but also generates redundant output as every node pair of the two input graphs must be considered.

\begin{figure}
\centering
    \includegraphics[width=\columnwidth]{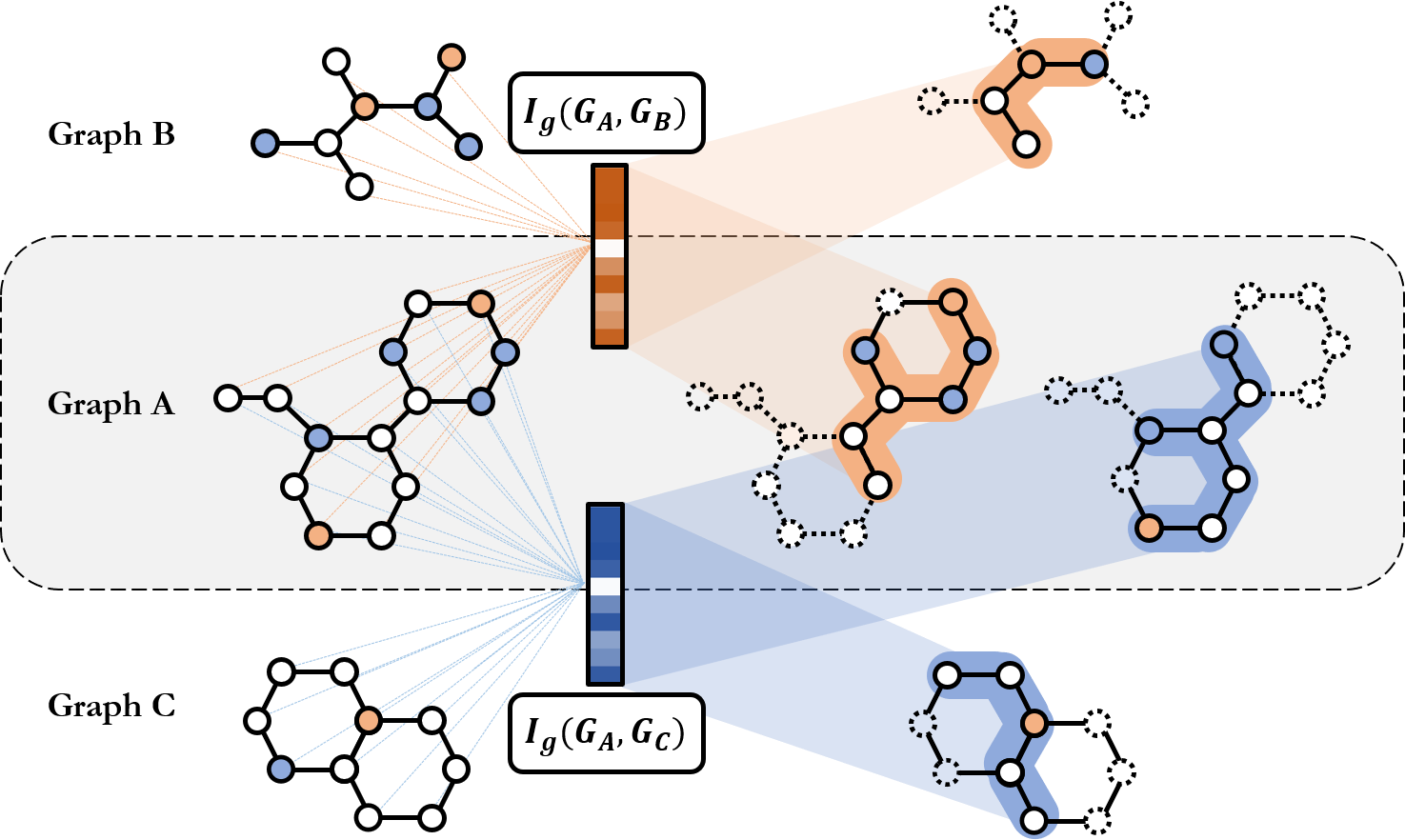}
    \caption{Illustration of the concept of co-attention graph pooling. The function $I_g(\cdot, \cdot)$ represents the interaction representation between two graphs, as described in Section \ref{sec:method}. Our proposed pooling method is able to extract different sub-graphs from the same graph representation when considering different pairs. Graph A has representations from different sub-graphs when paired with Graph B and Graph C.}
    \label{fig:fig_illustration}
\end{figure}

In this paper, we propose an efficient method for considering interactions between graphs at the \textit{graph-level} by applying co-attention to graph pooling. 
Our Co-attention Graph Pooling (CAGPool) dynamically represents each input graph based on its interaction with the opposite graph in the pair (as illustrated in Figure \ref{fig:fig_illustration}), while adding minimal computation complexity. 
Our model outperforms baselines even without using the additional information commonly used in baseline methods on real-world public benchmark datasets for both classification and regression tasks in paired graph representation learning: drug-drug interaction classification and graph similarity regression.
The implementation is publicly available\footnote{https://github.com/LeeJunHyun/CoAttentionGraphPooling}.

\section{Related Work}
\label{sec:related_work}
The overall pipeline of our model is divided into two components: Graph Convolution and Graph Pooling with Co-attention.
In this section, we provide a detailed overview of the previous works related to each component.

\subsection{Graph Convolution Networks}  
\label{subsec:GCN}
Researchers have been actively working on using convolutional neural networks to process graph-structured data.
However, the conventional convolution operation, which is defined on grids, is difficult to be directly applied on graphs due to irregular structure.
To address this issue, previous researches have defined graph convolution by using the Fourier transform (i.e. spectral graph convolution) \cite{kipf2017semi,monti2017geometric} or spatial graph connectivity (i.e. non-spectral graph convolution) \cite{hamilton2017inductive,velickovic2017graph}.
The spectral approaches can be represented by the work of Kipf and Welling \cite{kipf2017semi}, where graph convolution is redefined on the Fourier domain with a localized first-order approximation.
Non-spectral graph convolutions align to the work of Hamilton \cite{hamilton2017inductive},where graph convolution is defined as an aggregation function of a node's neighborhood representations.
Building upon these works, subsequent studies have aimed to improve the expressive power of both node-level and graph-level representations \cite{gcnii_icml20,pna_neurips20,magna_ijcai21}.

\subsection{Graph Pooling}
\label{subsec:pooling}
Pooling is a technique that prevents overfitting in modern neural network models with a large number of parameters by reducing the size of the representations.
This allows the model to generalize better to new data. There are two main categories of graph pooling methods: global pooling and hierarchical pooling.
Hierarchical pooling methods can be further grouped into \textit{pooling by graph transformation} and \textit{pooling by node selection}, based on how they produce the reduced node and edge sets.

\textbf{Global graph pooling} methods are techniques that transform the variable-sized representation of nodes produced by GNNs into a fixed-sized vector representation. This can then be used as input for downstream tasks such as prediction.
Some common global graph pooling methods include using simple aggregation functions such as summation, average, and max to combine the node representations.
The SortPool method \cite{zhang2018end} involves selecting the top $k$ nodes from the graph based on a certain criterion, and using their representations as the pooled output.
SimGNN \cite{bai2019simgnn} proposes using global context-aware attention to weight the contributions of different nodes, and producing a pooled representation as a weighted sum of all the node representations in the graph.
These methods allow for the use of GNN output as input for downstream tasks, and have been applied to a range of tasks including graph classification and regression.
\newline

\textbf{Pooling by graph transformation} is one of the hierarchical graph pooling methods where nodes are clustered and merged under certain criteria \cite{ying2018hierarchical,Yuan2020StructPool}.
The main purpose of \textit{Pooling by graph transformation} is to learn the assignment matrix that transforms input nodes into new cluster nodes.
The new adjacency matrix of cluster nodes is re-defined according to the assignment matrix.
The \textit{Pooling by graph transformation} has the advantage of keeping all node information. 
However, there is a computational complexity issue when calculating the new adjacency matrix for cluster nodes.
And also, the output of several pooling layers is not easily interpreted through the transformed graph.
\newline

\textbf{Pooling by node selection} creates hierarchical graph representations via trainable indexing method.
TopKPool \cite{gao2019graph} uses a trainable projection vector to calculate node scores and accordingly select nodes with the top-k score.
SAGPool \cite{pmlr-v97-lee19c} improves upon TopKPool by using GNNs to consider the topology of the graph along with the node features to calculate node scores.
Our method aligns with the work of \textit{pooling by node selection} as it has comparably lower complexity than pooling by graph transformation.
Although information loss might occur during node discarding, we propose that this helps the model to dynamically focus on different sub-graphs for different graph pairs by eliminating the representations of the nodes that are irrelevant to the interaction.

\subsection{Co-Attention Mechanism}
\label{subsec:attention}

Attention mechanisms, which allow neural networks to assign trainable importance weights to input~\cite{vaswani2017attention}, have been widely used in deep learning research.
There have been several works that extend the use of attention to input in the form of pairs. For example, Seo et al. \cite{seo2016bidirectional} proposed a bidirectional attention flow model that includes both query-to-context and context-to-query attentions (i.e., co-attention) for machine comprehension tasks.
Additionally, Deac et al. \cite{deac2020empowering} demonstrated that co-attention obtained from pairs of graphs can significantly improve the predictive power of GNNs.
In this work, we leverage the co-attention mechanism for graph pooling to select nodes in paired graphs that should be aware of each other.

\section{Methods}
\label{sec:method}
\begin{figure*}
    \centering
    \includegraphics[width=\textwidth]{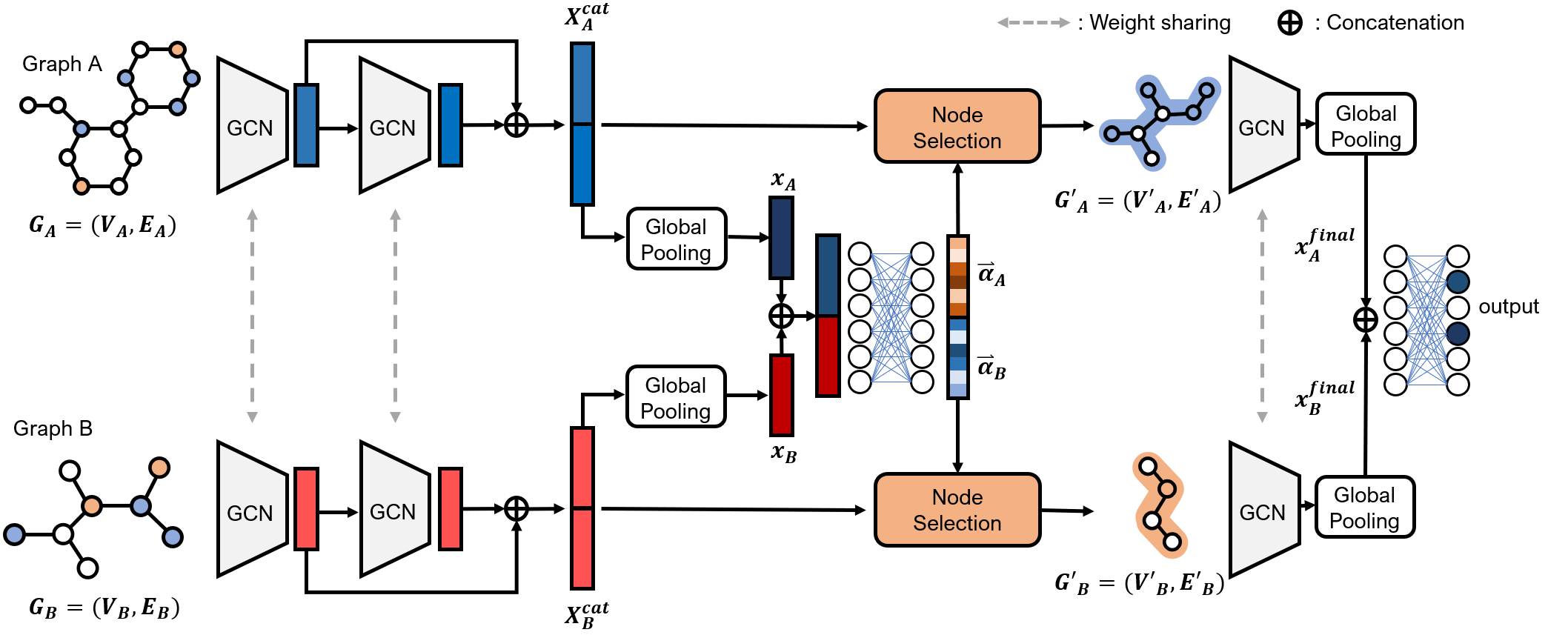}
    \caption{The overall architecture of GNN with Co-Attention Graph Pooling.}
    \label{fig:fig_pipeline}
\end{figure*}

In this section, we first describe the basic notations for general pairwise graph representation learning.
Then, we introduce our proposed method: Co-Attention Graph Pooling.
We denote the goal of our Co-Attention Graph Pooling, i.e., CAGPool, followed by detailed explanations for each components of our overall pipeline for pair-wise graph representation learning using CAGPool (see Figure~\ref{fig:fig_pipeline} for detailed illustration).

\subsection{Problem Setting}
Let $G=(V,E)$ denote the undirected graph, where $V$ is the set of vertices and $E$ is the set of edges.
Its adjacency matrix $A$ can be constructed with $A_{i,j}=1$ if $(i,j) \in E$ and $A_{i,j} = 0$ otherwise.
Node attributes can be represented as a matrix form: $X^{(0)} \in \mathbb{R}^{N \times F}$, where $N=|V|$ is the number of nodes and $F$ is the feature dimension.
Our main task is to predict the labels for a pair of input graphs $(G_\mathcal{A}, G_\mathcal{B})$, thus designing a model that learns the function $f: G \times G \mapsto R \in \mathbb{R}^{o}$, where $o$ denotes the output dimension.

While learning $f$, it has been studied that considering the interaction between the input graph pairs leads to better prediction~\cite{bai2019simgnn,deac2020empowering}.
For clear comparison with previous methods, we define $I(G)$ as the representation of the graph $G$ and the interaction representation of two graphs as $I(G_\mathcal{A},G_\mathcal{B})$, where $I_n(G_\mathcal{A},G_\mathcal{B})$ and $I_g(G_\mathcal{A},G_\mathcal{B})$ each denotes the interaction at the node-level and graph-level, respectively.
Since interactions were utilized at the node-level ($I_n(G_\mathcal{A}, G_\mathcal{B})$) in previous works~\cite{bai2019simgnn,deac2020empowering,pathak2020chemically}, they suffer from high complexity (see Figure \ref{fig:fig_complexity} and  Section \ref{sec:complexity}).

The main goal of this paper is to consider the interaction representation at graph-level (i.e., $I_g(G_\mathcal{A}, G_\mathcal{B})$) instead of node-level (i.e., $I_n(G_\mathcal{A}, G_\mathcal{B})$).
To this end, we propose a graph pooling module that learns a mapping function $g: G \times G \mapsto G' \times G'$ that improves the predictive power of $f$, where $G'=(V',E')$ is the sub-graph with $V' \subset V$ and $E' \subset E$.
The individual representations $I(G_\mathcal{A}), I(G_\mathcal{B})$ for each graph are then refined to $I(G_\mathcal{A}'), I(G_\mathcal{B}')$ respectively by using the interaction representation $I_g(G_\mathcal{A},G_\mathcal{B})$.

\subsection{Node Embedding with Graph Convolution}
\label{subsec:node_embedding}
Our overall pipeline follows the basic architecture of graph convolution Siamese network\cite{koch2015siamese}.
We receive a pair of graphs as input and apply the graph convolution with shared weights to each graph of the pair.
In the graph convolution layer, we update node representations by neighborhood aggregation.
We use the graph convolution suggested by Kipf and Welling \cite{kipf2017semi},
\begin{equation}
    X^{(l+1)}=\sigma(\tilde{D}^{-\frac{1}{2}}\tilde{A}\tilde{D}^{-\frac{1}{2}}X^{(l)}\Theta^{(l)})
\end{equation}{}
where $X^{(l)}$ is the node representation of $l$-th layer.
$\tilde{A}\in\mathbb{R}^{N\times N}$ is the adjacency matrix with self-loops.
$\tilde{D}\in \mathbb{R}^{N \times N}$ denotes the degree matrix of $\tilde{A}$. $\Theta^{(0)}\in\mathbb{R}^{F\times F'}$ and $\Theta^{(l)}\in\mathbb{R}^{F'\times F'} (l>=1)$ are learnable convolution weights.
$\sigma$ is a non-linear function that follows each graph convolution layer.
Afterwards, we concatenate the output of each output layer.
The final node representation $X^{cat} \in\mathbb{R}^{N\times nF'}$ is obtained by concatenating each of the $n$ convolution blocks.
\begin{equation}
    X^{cat}=X^{(1)}\|X^{(2)}\|...\|X^{(n)}
    \label{eq:X_cat}
\end{equation}{}
where $\|$ denotes the concatenation.
Because we have two graphs as a pairwise input, two node representations $(X^{cat}_\mathcal{A},X^{cat}_\mathcal{B})$ are obtained by a graph convolution Siamese network.

\subsection{Co-Attention Graph Pooling}
After individually obtaining the representations for the pair of input graphs, our CAGPool layer takes the two graph representations ($(X_{\mathcal{A}}^{cat},A_{\mathcal{A}})$ and $(X_{\mathcal{B}}^{cat},A_{\mathcal{B}}))$ after several graph convolution layers and then constructs the co-attention vector $\vec{\alpha}$ to calculate node scores.
Then, the subgraphs are extracted by indexing each graph with the node scores.
Below, we cover details of how CAGPool works.\newline

\textbf{Obtaining co-attention vector.}
The co-attention vector $\vec{\alpha}$ is obtained from the two graph-level representations of the input graph pair ($x_\mathcal{A},x_\mathcal{B}$), where each graph-level representation $x\in\mathbb{R}^{nF'}$ is obtained by global graph pooling.
In this paper, we chose global mean pooling as

\begin{equation}
\label{eq:gap}
    x=\frac{1}{N}\sum_{r=1}^{N}X^{cat}_r,
\end{equation}{}
where $X^{cat}_r \in \mathbb{R}^{nF'}$ is a $r$-th row vector of node representation matrix $X^{cat} \in \mathbb{R}^{N \times nF'}$.
Other global graph pooling methods can also be applied.

The graph-level representations are then concatenated and fed into a linear transformation layer to extract the co-attention vector as
\begin{equation}
\label{eq:alpha}
    \vec{\alpha}=W_\alpha[x_\mathcal{A}\|x_\mathcal{B}]+b_\alpha,
\end{equation}{}
where $W_\alpha$ and $b_\alpha$ are both trainable parameters with dimension of $W_\alpha\in\mathbb{R}^{2nF'\times2nF'}$ and $b_\alpha\in\mathbb{R}^{2nF'}$.
The extracted co-attention vector is then indexed as $\vec{\alpha}_\mathcal{A}=\vec{\alpha}_{0:nF'}\in\mathbb{R}^{nF'}$ and $\vec{\alpha}_\mathcal{B}=\vec{\alpha}_{nF':2nF'}\in\mathbb{R}^{nF'}$.
The multilayer perceptron (MLP) can also be used instead of a single linear transformation.
\newline

\textbf{Node selection using co-attention vector.}
The score of $r$-th node $Z_r\in\mathbb{R}$ is calculated by dot product of $X^{cat}_r\in\mathbb{R}^{nF'}$ and $\vec{\alpha}\in\mathbb{R}^{nF'}$.
All node scores $Z\in\mathbb{R}^{N}$ can be calculated as
\begin{equation}
    Z=\frac{X^{cat}\cdot \vec{\alpha}}{\|\vec{\alpha}\|},\quad\mbox{idx}=\mbox{TopK}(Z, \lceil kN \rceil )
\end{equation}
where $\cdot$ denotes dot product, TopK($\cdot$) function returns the indices of top $\lceil kN \rceil$ nodes according to $Z$, and $k \in (0,1]$ is a pooling ratio. 
We hold $\mbox{idx}_\mathcal{A}$ from $Z_\mathcal{A}$ and $\mbox{idx}_\mathcal{B}$ from $Z_\mathcal{B}$ for graphs $G_\mathcal{A}$ and $G_\mathcal{B}$, respectively.

Then, following the procedure of \textit{pooling by node selection} methods, we treat each node score as the significance of the corresponding node and select the top $\lceil kN \rceil$ nodes by indexing with $\mbox{idx}$ as
\begin{equation}
    X'=X_{\mbox{idx},:}^{cat} \odot Z_{\mbox{idx}}, \quad A'=A_{\mbox{idx,idx}},
    \label{eq:idx}
\end{equation}
where $(\cdot)_{\mbox{idx}}$ denotes the indexing operation and $\odot$ is the broadcasted elementwise product. $X'\in\mathbb{R}^{\lceil kN \rceil \times nF'}$ and $A' \in \mathbb{R}^{\lceil kN \rceil \times \lceil kN \rceil}$ are the node feature matrix and the adjacency matrix of a pooled graph $G'=(V',E')$, respectively.
Since $\vec{\alpha}$ is obtained from both $G_\mathcal{A}$ and $G_\mathcal{B}$, sub-graphs are extracted according to the interaction representation earlier denoted as $I_g(G_\mathcal{A},G_\mathcal{B})$.
Now we have $(X_{\mathcal{A}}',A_{\mathcal{A}}')$ for sub-graph $G'_{\mathcal{A}}$ and $(X_{\mathcal{B}}',A_{\mathcal{B}}')$ for sub-graph $G'_{\mathcal{B}}$.
\newline

\textbf{Prediction using sub-graphs.}
The sub-graphs obtained by Equation~(\ref{eq:idx}) are embedded again with graph convolution as
\begin{equation}
\label{eq:subnode}
    X'^{(l+1)}=\sigma(\tilde{D'}^{-\frac{1}{2}}\tilde{A'}\tilde{D'}^{-\frac{1}{2}}X'^{(l)}\Theta'^{(l)}).
\end{equation}{}
To feed to the final MLP layer, we convert the arbitrary sized representations from Equation~(\ref{eq:subnode}) into fixed-size vectors $x_\mathcal{A}^{final}, x_\mathcal{B}^{final}$  using the global graph pooling described in Equation (\ref{eq:gap}).
Then we concatenate the two graph-level representations and feed to the MLP layers as
\begin{equation}
    \mbox{output}=\mbox{MLP}([x_\mathcal{A}^{final} \| x_\mathcal{B}^{final}]).
\end{equation}{}
Depending on the task, appropriate functions (e.g. sigmoid, softmax) are applied to the final output.
\section{Experiments}
\label{sec:experiment}
We evaluate our method under two tasks: pairwise graph classification and pairwise graph regression.
In this section, we first outline the experimental settings and describe the datasets used to evaluate each task. 
Following that, we quantitatively compare our proposed CAGPool with the baseline methods to validate its effectiveness. 
During the training process, we use the Adam Optimizer with a learning rate of 1e-3 for 20 epochs and a batch size of 32. We fix the hidden dimension $F'$ via grid search across 32, 64, 128, and 256. 
All models, including the baselines, were implemented using PyTorch \cite{paszke2019pytorch} and PyTorch Geometric \cite{fey2019fast}. 
We carried out all experiments on a single NVIDIA Titan Xp GPU. 
The reported performance represents an average of the results from five repetitions.

\subsection{Tasks and Datasets}
\begin{table}
  \caption{AUROC results on the Decagon test set. The \checkmark under the feature+ column indicates that the according baseline leverages external features other than the drug itself such as protein-protein interaction or the one-hot encoding of the drug-drug interaction (the target class).
}
  \centering
  \footnotesize{
  \begin{tabular}{l|c|c}
    \textbf{Methods} & \textbf{feature+} & \textbf{AUROC} \\ \hline \hline
    Concatenated features & \checkmark & 0.793 \\
    Decagon & \checkmark & 0.872 \\
    Late-Outer & \checkmark & 0.724 \\
    CADDI& \checkmark & 0.778 \\
    MHCADDI & \checkmark & 0.882 \\ \hline 
    MPNN-Concat  & - & 0.661 \\
    Drug-Fingerprints  & - & 0.744 \\
    RESCAL& - & 0.693 \\
    DEDICOM & - & 0.705 \\
    DeepWalk & - & 0.761 \\
    MHCADDI-ML & - & 0.819 \\ \hline
    \textbf{Ours} & - & \textbf{0.891} \\
  \end{tabular}
  }
  \label{tab:AUROC}
\end{table}

\textbf{Drug-drug interaction prediction.}
Polypharmacy refers to the simultaneous use of multiple drugs by a single patient to treat one or more conditions.
One of the key challenges with polypharmacy is that patients may experience unexpected side-effects when they take several drugs concurrently. 
These side-effects can lead to potentially devastating clinical and financial outcomes, including post-marketing withdrawal of drugs from the market \cite{onakpoya2016post}. 
Consequently, it is of paramount importance to accurately predict potential DDIs of drug candidates during the drug discovery process \cite{han2017synergistic}.

We utilized DDI dataset collected and used by Decagon \cite{zitnik2018modeling}. 
Decagon adopts a pathway-based approach, integrating Protein-Protein, Drug-Protein, and Drug-Drug interaction networks into a singular graph structure. 
In this structure, each node represents a drug or protein, while the links indicate the interactions between these entities. 
Following the experimental setup of \cite{deac2020empowering}, we exclusively used the Drug-Drug Interaction subset of this dataset. 
In our representation, each drug is depicted as an individual graph with atoms serving as nodes and chemical bonds as edges. 
The structural information for each drug compound is obtained using Rdkit\footnote{http://www.rdkit.org}. 
The ultimate goal of our experiment is to predict all types of DDI (if any) that might occur, using solely the structural information of two given drug compounds.


\begin{table}
  \caption{Evaluation results under AUPRC and AP@50 on the Decagon test set for a more precise comparison with baselines that provide their performance under these metrics.}
  \centering
  \small{
  \begin{tabular}{l|ccc}
    \textbf{Methods} & \textbf{AUROC} & \textbf{AUPRC} & \textbf{AP@50} \\ \hline  \hline
    RESCAL & 0.693 & 0.613 & 0.476 \\
    DEDICOM & 0.705 & 0.637 & 0.567 \\
    DeepWalk & 0.761 & 0.737 & 0.658 \\
    Concatenated features & 0.793 & 0.764 & 0.712 \\
    Decagon & 0.872 & 0.832 & 0.803 \\ \hline
    \textbf{Ours} & \textbf{0.891} & \textbf{0.867} & \textbf{0.812} \\
  \end{tabular}}
  \label{tab:eval_table}
\end{table}

We adopt the exact filtering steps employed by the baselines \cite{zitnik2018modeling,deac2020empowering} on the Decagon dataset, including negative sampling. 
A detailed explanation of the negative sampling process is provided in our supplementary material. 
We use the 964 types of polypharmacy side effects that occur more than 500 times. 
The complete dataset consists of 4,576,785 positive examples. 
We allocate 80\% of the interactions to the training set, 10\% to the validation set, and the remaining 10\% to the test set. 
The evaluation results are reported on the test set, for the model that achieved the best performance on the validation set. 
During the testing phase, we calculate the AUROC, AUPRC, and AP@50 across the 964 drug-drug interaction classes, exclusively for valid samples (either positive samples or those obtained via negative mining).
\newline



\begin{table*}
  \caption{The experimental results of graph similarity regression task. The evaluation metrics are Mean Square Error (MSE, $10^{-3}$), Spearman’s rank correlation coefficient $\rho$, and Kendall’s rank correlation coefficient $\tau$.} 
  \centering
  \begin{tabular}{lc|ccc|ccc|ccc}
    && \multicolumn{3}{c|}{\textbf{AIDS}} & \multicolumn{3}{c|}{\textbf{LINUX}} & \multicolumn{3}{c}{\textbf{IMDB-M}}  \\
    \textbf{Methods}    & \textbf{Complexity} & MSE $(\downarrow)$ & $\rho$ $(\uparrow)$& $\tau$ $(\uparrow)$& MSE $(\downarrow)$ & $\rho$ $(\uparrow)$& $\tau$ $(\uparrow)$& MSE $(\downarrow)$ & $\rho$ $(\uparrow)$ & $\tau$ $(\uparrow)$\\
    \hline \hline
    SimGNN & $O( \max(|V_A|,|V_B|)^2)$ & \textbf{1.189}  & 0.843 & 0.690 & 1.509  & 0.939 & \textbf{0.830} & 1.264  & 0.878 & 0.770 \\
    GraphSim & $O(|V_A||V_B|)$ & 1.919   & 0.849 & - &  0.471 & 0.976 & - & 0.743  & 0.926 & -    \\
    GMN & $O(|V_A||V_B|)$ &  1.886  & 0.751 & - &  1.027 &0.933 & - &  4.422 & 0.725 & -    \\
    MPNGMN & $O(|V_A||V_B|)$ &  1.191  & 0.904 & \textbf{0.749} &  1.561 & 0.945 & 0.814 & 1.331  & 0.889 & -    \\
    GENN & $O(|V_A||V_B|)$ &  1.618  &  0.901 & - &  0.438 & 0.955 & - &  0.883 & 0.880 & -    \\
    GED-CDA & $O(|V_A||V_B|+|V_A|^2+|V_B|^2)$ &  1.367  & \textbf{0.914} & - &  \textbf{0.125} & \textbf{0.985} & - &  \textbf{0.711} & \textbf{0.919} & -    \\
    \hline
    S-Mean & $O(|E_A|+|E_B|)$ & 3.115  & 0.633 & 0.480 & 16.950  & 0.020 & 0.016 & 3.749  & 0.774 & 0.644     \\
    H-Mean & $O(|E_A|+|E_B|)$ & 3.046  & 0.681 & 0.629 & 6.431  & 0.430 & 0.525 & 5.019  & 0.456 & 0.378     \\
    H-Max & $O(|E_A|+|E_B|)$  &3.396 & 0.655 & 0.505  & 6.575 &0.879& 0.740 & 6.993& 0.455 &0.354     \\
    Att.Deg.  & $O(|E_A|+|E_B|)$&   3.338 & 0.628 & 0.478 &8.064 &0.742 &0.609 &2.144 &0.828 &0.695 \\
    Att.GC  & $O(|E_A|+|E_B|)$& 1.472&  0.813&  0.653& 3.125&  0.904&  0.781&  3.555 & 0.684 & 0.553\\
    Att.LGC  & $O(|E_A|+|E_B|)$&  \textbf{1.340} &  0.825 &  0.667 & 2.055 &  0.916 &  0.804 &   1.455 &  0.835 &  0.700 \\
    SGNN & $O(|E_A|+|E_B|)$ &  2.822  & 0.765 & 0.588 & 11.832  & 0.566 & 0.404 &  1.430 & 0.870 & -    \\
    CAGPool (Ours) & $O(|E_A|+|E_B|)$ &  2.959 &  \textbf{0.880} &  \textbf{0.717} &\textbf{0.658} &\textbf{0.988} &\textbf{0.922} & \textbf{0.884} &\textbf{0.975} &\textbf{0.874}    \\
    \hline
  \end{tabular}
  \label{tab:GED_results}
\end{table*}

\textbf{Graph similarity prediction.}
Graph retrieval is a fundamental problem which involves calculating the distance or similarity between two graphs. 
The Graph Edit Distance (GED) is the most widely used distance metric for graph retrieval \cite{bai2019simgnn,bunke1983distance,ktena2017distance,liang2017similarity,zhao2013partition,zheng2013graph}. 
The edit distance is defined as the minimum number of operations needed to transform $G_\mathcal{A}$ into $G_\mathcal{B}$. 
However, computing the GED is known to be an NP-complete problem \cite{bunke1998graph}, making it infeasible to calculate the exact GED within a reasonable time frame for graphs that have more than 16 nodes \cite{blumenthal2018exact}.

Recently, there are attempts to approximate GED by using GNNs \cite{bai2019simgnn}.
The authors also provide GED datasets containing AIDS, LINUX, and IMDB \footnote{https://github.com/yunshengb/SimGNN}.
\textit{AIDS} \footnote{https://wiki.nci.nih.gov/display/NCIDTPdata} is commonly used in graph similarity search \cite{ktena2017distance,liang2017similarity,zhao2013partition,zheng2013graph}. 
AIDS dataset contains chemical compound structure graphs with labeled nodes.
Bai et al. \cite{bai2019simgnn} selected 700 graphs of equal or less than 10 nodes each. 
\textit{LINUX} \cite{wang2012efficient} dataset is about program dependence graphs generated from the Linux kernel.
In the program function graphs of the LINUX dataset, a node represents a statement and an edge is a dependency between two statements.
Bai et al. \cite{bai2019simgnn} selected 1000 graphs, each of which has equal or less than 10 nodes.
For both AIDS and LINUX datasets, the ground truth GEDs are calculated by using the $A^*$ algorithm.
\textit{IMDB} \cite{yanardag2015deep} dataset contains 1500 graphs of which the nodes represent actors or actresses.
Edges denote that two people appear in the same movie.
As in SimGNN, we use all the graphs in IMDB dataset for testing the scalability.
For the IMDB dataset, the approximation algorithms, Beam \cite{neuhaus2006fast}, Hungarian \cite{kuhn1955hungarian}, and VJ \cite{fankhauser2011speeding}, were used for the ground truth because the IMDB dataset contains graphs with more than 16 nodes.
GED is converted to similarity score $S$ with normalization ($nGED$) as follow: $nGED(G_\mathcal{A},G_\mathcal{B}) = \frac{GED(G_\mathcal{A},G_\mathcal{B})}{(|G_\mathcal{A}|+|G_\mathcal{B}|)/2}, S(GED(G_\mathcal{A},G_\mathcal{B})) =\exp^{-nGED(GED(G_\mathcal{A},G_\mathcal{B}))}$ ,where $|G_i|$ denotes the number of nodes of graph $G_i$ and the similarity score $S$ is in the range of (0,1].

We exactly follow their training/testing data split and randomly select validation set within a training set with the same proportion.
All the graphs are split into 60\%, 20\%, and 20\% as a training set, a validation set, and a testing set. 

\subsection{Evaluation and Baselines}
\textbf{Drug-drug interaction prediction.}
In the DDI prediction task, we assess the effectiveness of CAGPool by comparing it with existing methods using the Decagon dataset. 
We primarily employ three evaluation metrics used in Decagon~\cite{zitnik2018modeling}: AUROC, AUPRC, and AP@50.

Table \ref{tab:AUROC} presents a comparison of our network with existing baseline networks on the Decagon dataset. 
The \checkmark in the \textbf{feature+} column indicates that the method uses additional information beyond the structural information of drugs. 
The \textbf{Concatenated features} method~\cite{zitnik2018modeling} employs a PCA representation of the drug-target protein interaction matrix and individual drug side-effects. 
The \textbf{Decagon} method~\cite{zitnik2018modeling} further includes protein-protein interactions, drug-protein target interactions, and single drug side-effect information. Methods such as \textbf{MPNN-Concat}, \textbf{Late-Outer}, \textbf{CADDI}, \textbf{MHCADDI}~\cite{deac2020empowering} use one-hot encoding of interactions to predict their presence or absence. 
Our model outperforms all these methods, whether they use additional information or not, by only leveraging the structural features of the graph representation for drugs.

Table \ref{tab:eval_table} presents a performance comparison across AUROC, AUPRC, and AP@50 metrics on the Decagon dataset. 
Not only does our method outperform the baseline methods proposed in \cite{zitnik2018modeling}, but it also demonstrates consistently superior performance across all evaluation metrics. 
Both \textbf{RESCAL} and \textbf{DEDICOM} are tensor decomposition approaches applied to the drug-drug matrix, while \textbf{DeepWalk} employs neural embedding based on a random walk procedure.
\newline

\textbf{Graph similarity prediction.}

Table \ref{tab:GED_results} presents the regression performance on GED datasets. 
The baselines encompass GNN approaches reported in the GED-CDA paper \cite{10094975} and the SimGNN paper \cite{bai2019simgnn}. \textbf{SimGNN} \cite{bai2019simgnn}, \textbf{GraphSim} \cite{bai2020learning}, \textbf{GMN} \cite{li2019graph}, \textbf{MPNGMN} \cite{ling2020hierarchical}, \textbf{GENN} \cite{wang2021combinatorial}, and \textbf{GED-CDA} \cite{10094975} are characterized by their focus on node-wise interactions.
This results in a substantial computational burden due to their high complexity, which is at least $O(|V_A||V_B|)$.
For instance, \textbf{SimGNN} combines \textit{AttLearnableGC} and \textit{Pairwise Node Comparison}. 
To account for graph-graph interactions, \textit{SimGNN} utilizes the histogram information of the dot product of all node pairs between two graphs, an approach known as \textit{Pairwise Node Comparison}.
\textbf{SimpleMean (S-Mean)} generates a graph-level embedding by averaging the node representations. 
Both \textbf{HierarchicalMean (H-Mean)} and \textbf{HierarchicalMax (H-Max)} employ a graph coarsening algorithm for hierarchical graph representations \cite{defferrard2016convolutional}, and then apply global mean and max pooling, respectively.
In \textbf{AttDegree (Att.Deg.)}, the attention weight of nodes is calculated using the natural log. 
Both \textbf{AttGlobalContext (Att.GC)} and \textbf{AttLearnableGC (Att.LGC)} compute the attention weights using the graph-level representations. 
However, the latter also incorporates a learnable non-linear transformation, unlike the former.
SGNN \cite{ling2020hierarchical}, a Siamese architecture with GCN, has its performance reported in the GED-CDA paper. 
For a fair comparison, we maintain the model architecture of SimGNN and only replace its \textit{Pairwise Node Comparison} module with CAGPool.

\section{Discussion}
\label{sec:discussion}

\begin{figure*}
    \centering
    \includegraphics[width=\textwidth]{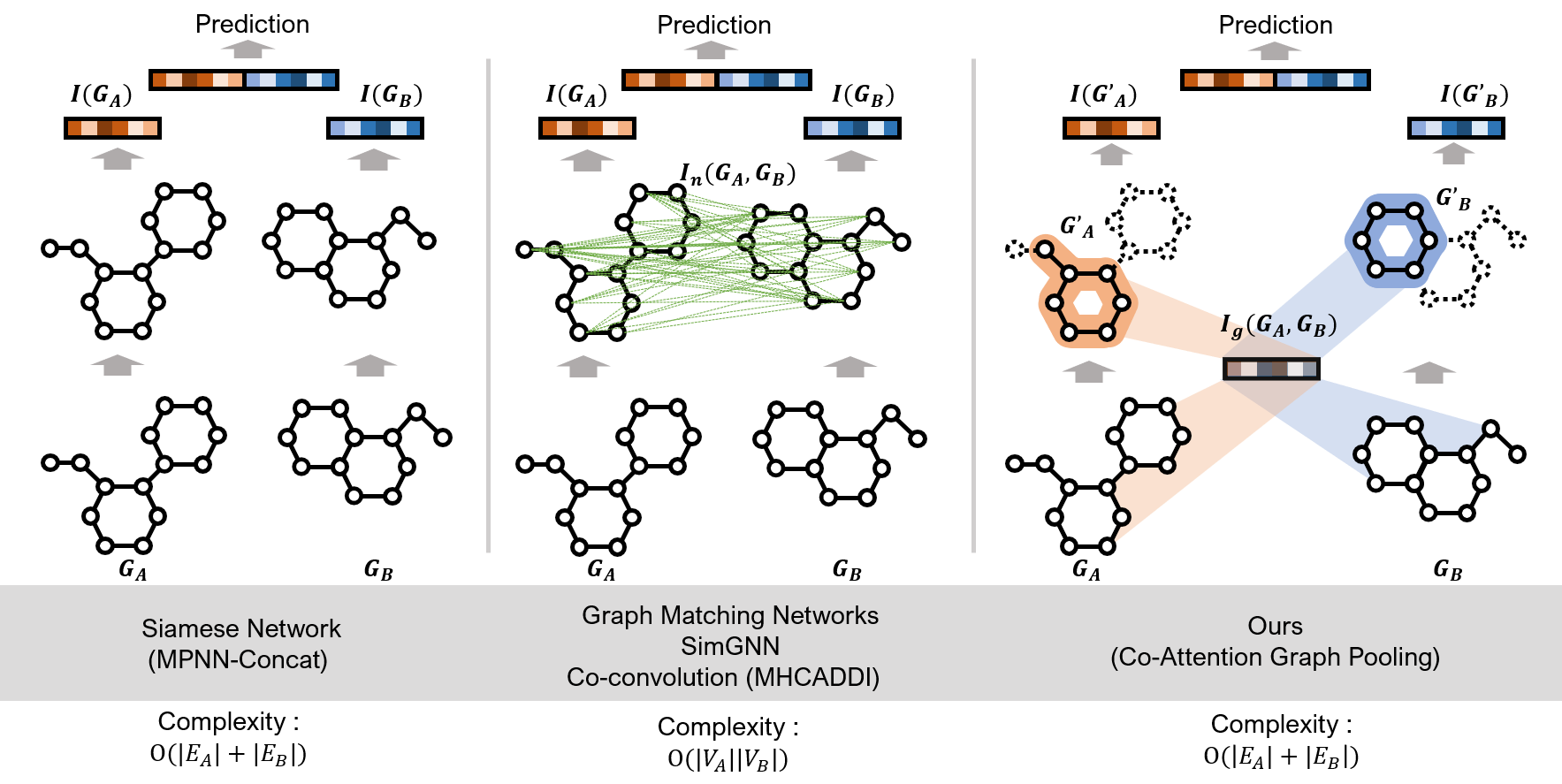}
    \caption{Comparison of the complexity between GNNs for pairwise graph inputs. 
    Given a pair of graphs $G_\mathcal{A}$ and $G_\mathcal{B}$, the graph Siamese network (left) has complexity $O(|E_\mathcal{A}| + |E_\mathcal{B}|)$, but cannot embed the graph-graph interaction. 
    Although SimGNN and MHCADDI (middle) can consider the interaction $I_n(G_\mathcal{A},G_\mathcal{B})$ in the node-level, the complexity increases to $O(|V_\mathcal{A}||V_\mathcal{B}|)$. 
    CAGPool (right) is able to consider the interaction representation $I_g(G_\mathcal{A},G_\mathcal{B})$ in the graph-level while maintaining the complexity upper bound of the graph Siamese network.
    In our notations, $I(G_\mathcal{A}'), I(G_\mathcal{B}')$, and $I_g(G_\mathcal{A},G_\mathcal{B})$ are aligned with vectors $x^{final}_\mathcal{A},x^{final}_\mathcal{B}$, and $\vec{\alpha}$, respectively.} 
    \label{fig:fig_complexity}
\end{figure*}
\subsection{Effectiveness of Considering the Interaction Representation}

The Graph convolution Siamese network~\cite{liang2017similarity} provides a basic framework to deal with pairwise graph inputs by predicting the interaction from individually embedded inputs (i.e., $I(G_\mathcal{A})$ and $I(G_\mathcal{B})$).
However, as the interaction is only minimally reflected in the final prediction layer, previous works have extended the scope to leverage interaction representations (i.e., $I_n(G_\mathcal{A},G_\mathcal{B})$) during the embedding stage.
\textit{MHCADDI} exchanges node representations between two graphs in the message-passing step, while \textit{SimGNN} compares all node pairs of two graphs after the node embedding layers.
Since \textit{MHCADDI} and \textit{SimGNN} show a substantial gain in predictive power compared to simple Siamese networks (see Table \ref{tab:AUROC} and Table \ref{tab:GED_results}), it can be concluded that utilizing $I(G_\mathcal{A},G_\mathcal{A})$ in the embedding stage is crucial when dealing with pairwise graphs.
However, as these methods utilize all node pairs, unnecessary complexity and redundant information are produced.
This results in suboptimal performance and efficiency, thereby compromising the final prediction.
Our CAGPool focuses on subgraphs that are extracted by the interaction representation, mitigating the issue of noisy representations.
The experimental results in Table \ref{tab:AUROC} and Table \ref{tab:GED_results} show that CAGPool successfully utilizes the interaction representations at the graph-level $I_g(G_\mathcal{A},G_\mathcal{A})$, achieving state-of-the-art performance on both tasks.

\subsection{The Efficiency of Co-attention Graph Pooling}
\label{sec:complexity}
Figure \ref{fig:fig_complexity} illustrates the complexity of each method that treats pairwise graph input.
The Graph convolution Siamese network, being the most basic form, has an overall complexity bounded by the complexity of embedding each graph with GNNs: $O(|E_\mathcal{A}| + |E_\mathcal{B}|)$.
Recent methods, such as \textit{MHCADDI} and \textit{SimGNN}, construct interaction representations at the node-level and require pairwise calculations for every node pair between $V_\mathcal{A}$ and $V_\mathcal{B}$.
Therefore, the computational complexity is bounded by $O(|V_\mathcal{A}||V_\mathcal{B}|)$.
Strictly speaking, \textit{SimGNN} has a complexity of $O(\mbox{max}(|V_\mathcal{A}|,|V_\mathcal{B}|)^2)$.
On the other hand, our approach requires additional computation with a complexity of $O(|V_\mathcal{A}| + |V_\mathcal{B}|)$ for constructing the interaction representation at the graph-level and selecting the nodes.
Therefore, CAGPool maintains the $O(|E_\mathcal{A}| + |E_\mathcal{B}|)$ complexity of the simple graph convolution Siamese network without increasing the complexity upper bound.
Additionally, we only need $W_\alpha \in \mathbb{R}^{2nF' \times 2nF'}$ and $b_\alpha \in \mathbb{R}^{2nF'}$ (see Equation (\ref{eq:alpha})) as additional trainable parameters, where $n$ is the number of GCN layers and $F'$ is the hidden dimension.
Given inputs $X_A$ and $X_B$, our module produces $X_A’$ and $X_B’$, demonstrating a 31.2 $\sim$ 64.7\% faster running time than the node-level interaction module when we set the number of nodes from 50 to 200.
We describe the details in the supplementary material.

\subsection{Ablation Study}
\begin{table}
\tiny
  \caption{Ablation studies and comparison with other pooling methods on the Decagon test set. Note that the scope of this ablation study only includes methods that do \textbf{not} use additional information such as protein features or one-hot encoding of the interaction.}
  \centering
  \small
  \begin{tabular}{l|ccc|c}
    \textbf{Methods} & \textbf{GNN} & \textbf{Co-att} & \textbf{Pooling} & \textbf{AUROC} \\ \hline \hline
    MPNN-Concat & \checkmark & - & - & 0.661 \\
    MHCADDI-ML & \checkmark & \checkmark & - & 0.819 \\ 
    TopKPool & \checkmark & - & \checkmark & 0.847 \\
    SAGPool  & \checkmark & - & \checkmark & 0.862 \\ \hline
    \textbf{Ours} & \checkmark & \checkmark & \checkmark & \textbf{0.891} \\
  \end{tabular}
  \label{tab:ablation}
\end{table}
We conduct an ablation study to validate (1) whether co-attention serves as an effective component when treating pairwise graphs, and (2) whether CAGPool demonstrates a meaningful improvement compared to individual pooling without considering interaction representation.

For (1), it can be concluded that leveraging co-attention is effective by comparing \textit{MHCADDI-ML} and CAGPool with \textit{MPNN-Concat} (the vanilla Siamese Network architecture in Figure~\ref{fig:fig_complexity}).
Moreover, CAGPool shows a significant improvement in performance even when compared to \textit{MHCADDI-ML}, implying that CAGPool is a more effective way to utilize the co-attention mechanism.

For (2), we only consider hierarchical pooling methods with node selection to match our experimental settings.
For a fair comparison with other hierarchical node selection-based pooling methods, we implemented TopKPool~\cite{gao2019graph} and SAGPool~\cite{pmlr-v97-lee19c} and kept the pooling ratio at 50\%.
Although all pooling methods show a gain in performance, our co-attention-based approach serves as the most effective pooling method for pairwise graph prediction.

\subsection{Future Works}

\textbf{Studies on the extracted subgraphs.}
The main focus of our work is the proposal of a novel pooling method, CAGPool, for extracting subgraphs from graph pairs, which we refer to as $G'_\mathcal{A} = (V'_\mathcal{A}, E'_\mathcal{A})$ and $G'_\mathcal{B} = (V'_\mathcal{B}, E'_\mathcal{B})$, that can help predict labels such as drug-drug interactions.
Even if some nodes, $v \in V'$, are isolated, they contain important information due to the use of GCN layers.
In the context of predicting drug-drug interactions, it can be extremely difficult to disambiguate the functional groups (i.e., subgraphs) related to a specific side effect, even for experts in the biomedical field.
This is because these subgraphs do not directly interact with each other, but rather affect each other through complex biological pathways within the human body.
While our research did not specifically investigate this analysis, we expect that our method can facilitate future research in this area by identifying subgraphs that are likely to be related to the functional groups responsible for the side effects between drugs.

\section{Conclusion}
\label{sec:conclusion}
In this paper, we describe the Co-Attention Graph Pooling (CAGPool) method for processing pairs of graph-structured data in an efficient and effective way. 
CAGPool combines the co-attention mechanism and pooling by node selection to enable a graph neural network to identify important sub-graphs for prediction tasks while maintaining the computational simplicity of Siamese networks. 
We demonstrate the effectiveness of our approach on two real-world benchmark datasets, showing that it outperforms baselines even without using additional information commonly utilized by other methods. 
We believe that CAGPool has the potential to be beneficial in a wide range of applications, including the urgent development of COVID-19 treatments, where the ability to identify key sub-graphs in paired data can help prevent unintended side effects.

\appendices

\section{\break Datasets}
\label{sec:datasets}
In this paper, we validate CAGPool on two tasks: a \textit{classification task} for Drug-Drug Interaction (DDI) prediction and a \textit{regression task} on the Graph Edit Distance (GED) dataset. 
Here, we provide additional information on the preprocessing details and analysis for each dataset.

\subsection{Drug-drug interaction dataset}
As described in the main text, we utilized the DDI dataset assembled by Decagon~\cite{zitnik2018modeling}. 
The original Decagon dataset contains Protein-Protein, Drug-Protein, and Drug-Drug pairs integrated into a graph structure. 
In this structure, each node represents either drugs or proteins, and the links indicate interactions between the two corresponding vertices. 
For our research, we focused solely on the drug-drug interaction subset. 
Each drug is represented as a single graph with atoms as nodes and chemical bonds as edges. 
The node attributes include the type of atoms, polarity, number of hydrogen atoms, and aromaticity. 
The original literature filtered for 500 or more drug pairs, but after our preprocessing, we found that to align with the 964 interaction types, we had to filter for 498 or more pairs.

\textbf{Negative Sampling}
To compensate for TWOSIDES, which only contains positive samples, we adopt the \textit{negative sampling} approach used in previous works~\cite{zitnik2018modeling,deac2020empowering}. 
The detailed process involving a drug $d$ and side-effect $se$ is described below:

\begin{itemize}
    \item During training, tuples ($\tilde{d_x},\tilde{d_y},se_z$), where $\tilde{d_x}$ and $se_z$ are chosen from the dataset and $\tilde{d_y}$ is chosen at random from the set of drugs difference from $d_y$ in the true samples ($\tilde{d_x},d_y,se_z$). The negative sample is selected randomly according to sampling distribution $P_r$, where for each node ${d_i}$ has a probability of $P(d_i)=\frac{f(d_i)^{3/4}}{\sum^{n}_{j=0}f(d_j)^{3/4}}$ of appearing \cite{mikolov2013distributed}.
    
    \item During validation and testing, we randomly sample two distinct drugs which do not appear in the positive dataset.
\end{itemize}

\subsection{Graph Edit Distance dataset}
We utilized the GED dataset from SimGNN\footnote{https://github.com/yunshengb/SimGNN}~\cite{bai2019simgnn}. 
The GED dataset comprises three sub-datasets: AIDS, LINUX, and IMDB. 
These datasets were designed and collected to evaluate the performance of the \textit{graph retrieval} task, a task designed to find similar/dissimilar graphs from the database when given a query graph.

\begin{figure}
    \centering
    \includegraphics[width=\columnwidth]{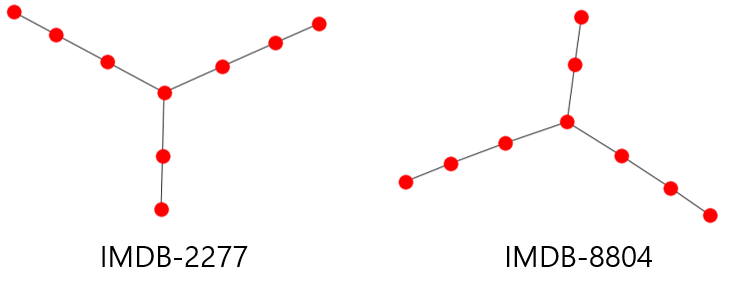}
    \caption{An example of isomorphic graphs.
    IMDB-2277 and IMDB-8804 are from different instance but have a same graph structure and node features.
    The graph edit distance between them is zero.
    }
    \label{fig:isomorphism}
\end{figure}

We found that some graphs in GED datasets were in an equivalence relation (i.e. isomorphism) (see Figure \ref{fig:isomorphism}).
According to SimGNN authors, they treat the query graph as an unseen graph even if there exists an isomorphic graph in the database.
Because checking for isomorphism is expensive with traditional graph algorithms, an algorithm that can efficiently capture isomorphic graphs is important in this dataset.
Since our model is permutation invariant, it can benefit this task by capturing isomorphism efficiently.

\section{\break Comprehensive Overview of Graph Pooling Methods}
\label{sec:overview}

\begin{figure*}
    \centering
    \includegraphics[width=0.866\textwidth]{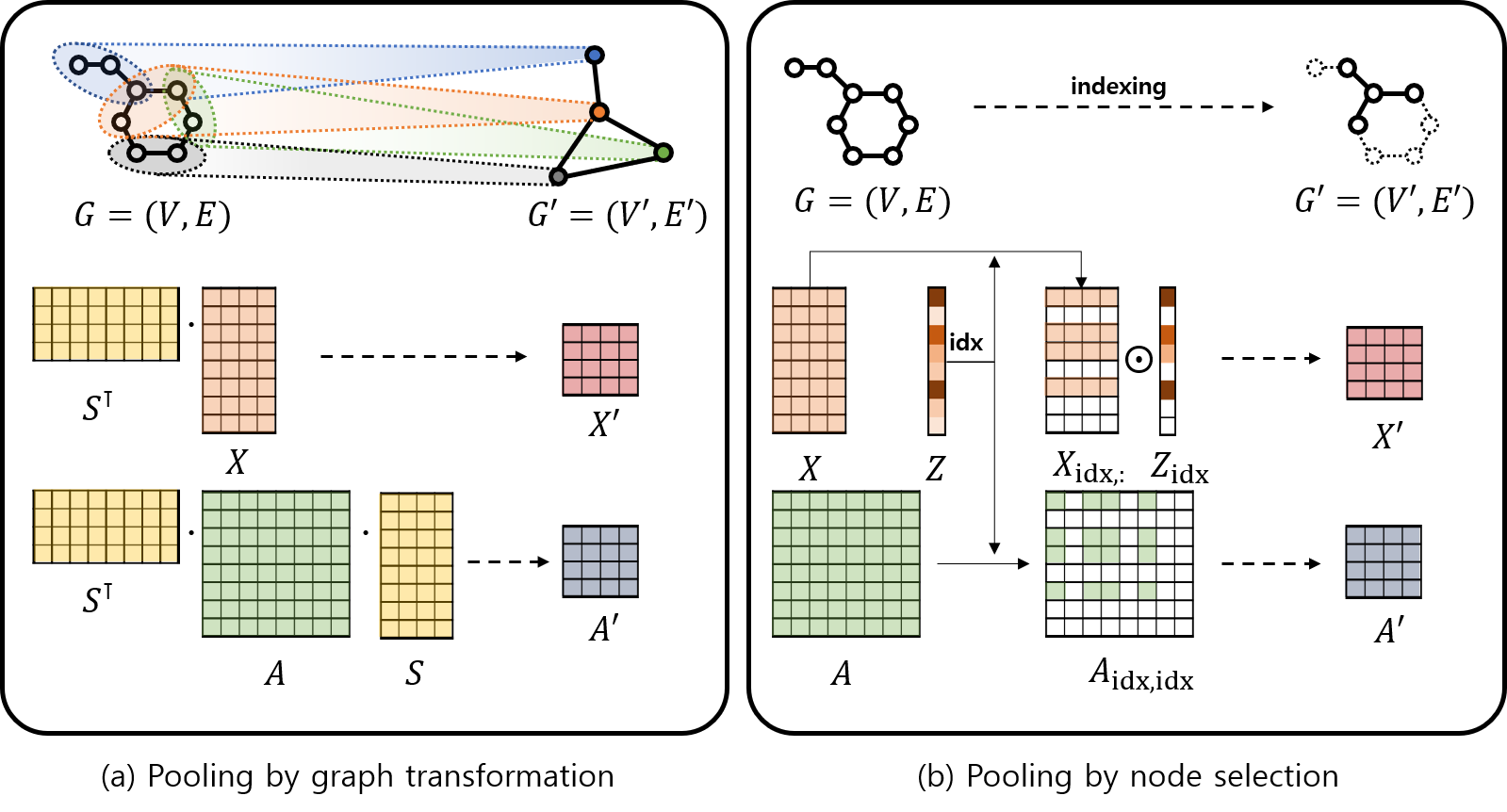}
    \caption{The illustration of \textit{pooling by graph pooling} and \textit{pooling by node selection}.
    $\cdot$ denotes the matrix multiplication and $\odot$ denotes the broadcasted multiplication.
    For both methods, the input graph $G=(V,E)$ has the node feature matrix $X \in \mathbb{R}^{N \times F}$ and the adjacency matrix $A \in \mathbb{R}^{N \times N}$, where $N,F$ denote the number of nodes and the feature dimension, respectively.
    In (a) pooling by graph transformation, the output graph $G' = (V',E')$ is generated according to the transformation matrix $S \in \mathbb{R}^{N \times N'}$, where $V'$ is the set of cluster nodes and $N'$ denotes the number of cluster nodes.
    In (b) pooling by node selection, it is important to define the node score $Z \in \mathbb{R}^{N}$ well.
    According to the node scores $Z$, the top-$k$ nodes are selected as elements of $V' \subset V$.
    }
    \label{fig:overview}
\end{figure*}

In this section, we 1) explain the categorization of graph pooling methods and 2) compare them in terms of complexity. 
Hierarchical graph pooling methods can be classified into \textit{pooling by graph transformation} and \textit{pooling by node selection}. 
The overview is illustrated in Figure \ref{fig:overview}. There are various pooling methods based on graph algorithms such as spectral clustering \cite{defferrard2016convolutional}, but we only consider graph pooling methods that can be trained end-to-end. 
Our approach is based on the \textit{pooling by node selection} method.

Given the input graph $G=(V,E)$ with the set of vertices $V$ and the set of edges $E$, the hierarchical graph pooling can be represented as the function $g: G \mapsto G'$, where $G'=(V',E')$ is the small-size graph.
Graph $G$ has the node feature matrix $X \in \mathbb{R}^{N \times F}$ and the adjacency matrix $A \in \mathbb{R}^{N \times N}$, where $N$ is the number of nodes and $F$ is the feature dimension.
The final goal of the graph pooling method is to obtain the graph $G'=(V',E')$ that has the node features $X' \in \mathbb{R}^{N' \times F}$ and the adjacency matrix $A' \in \mathbb{R}^{N' \times N'}$.
We categorize the graph pooling methods according to how $X'$ and $A'$ are calculated.

\subsection{Pooling by graph transformation}
\label{subsec:graph_transform}
Pooling by graph transformation (also called as pooling by clustering) downsamples graph by learning the transformation matrix.
Following the work of Ying et al. \cite{ying2018hierarchical}, pooling methods in this category calculate the output node features $X'$ and the adjacency matrix $A'$ by using the transformation matrix $S \in \mathbb{R}^{N \times N'}$, which is called as the assignment matrix~\cite{ying2018hierarchical}.
Here, the matrix $S$ transforms not only nodes, but also edges.
Therefore, how we define $S$ is the main key in \textit{pooling by graph transformation}.
In DiffPool, the transformation matrix $S$ is obtained from the node embedding of Graph Neural Networks (GNNs).
In StructPool, the transformation matrix $S$ is trained via conditional random fields \cite{Yuan2020StructPool:}.
Briefly, \textit{pooling by graph transformation} methods have a common framework as
\begin{equation}
X' = S^\top X, \quad A' = S^\top A S,
\label{eq:iter}
\end{equation}
where $X' \in \mathbb{R}^{N' \times F}$ is the feature matrix of $N'$ cluster nodes and $A' \in \mathbb{R}^{N' \times N'}$ is the adjacency matrix of them.
Note that $G'=(V',E')$ is not a sub-graph of $G$ (i.e. $V' \not \subset V, E' \not \subset E $).

\subsubsection{Complexity issue of graph transformation}
Despite the improvement in their performance, pooling by graph transformation suffers from heavy computational cost when obtaining the new adjacency matrix $A'$.
DiffPool suffers from heavy computational complexity because the $S$ and the output graph is represented as a dense matrix.
Although $S$ of StructPool can be either dense or sparse according to their setting, it still suffers from heavy complexity because of the iterative method for $S$ and the calculation of $A'$ described in Equation (\ref{eq:iter}).

\subsection{Pooling by node selection}
\label{subsec:node_sel}
In \textit{Pooling by node selection}, $X'$ and its adjacency matrix $A'$ are obtained by selecting the nodes according to the node score $Z \in \mathbb{R}^{N}$ , leaving nodes with high scores and discarding the rest.
Because both $X'$ and $A'$ are simply calculated by indexing, these methods do not increase the upper bound of computational complexity in GNNs.
Pooling by node selection therefore eventually boils down to ``how we define the scoring function for each node? (i.e., how we define $Z$?)"
TopKPool calculate node scores $Z$ from the dot product of node features and a trainable projection vector \cite{gao2019graph}.
SAGPool exploit both node features and the graph topology to calculate node scores $Z$ \cite{pmlr-v97-lee19c}.
After the calculation of $Z$, both perform the indexing operation as
\begin{equation}
 X' = X_{\mbox{idx},:}, \quad  A' =  A_{\mbox{idx},\mbox{idx}},
\end{equation}
where $(\cdot)_{\mbox{idx}}$ denotes the indexing operation and $\mbox{idx}$ is the top-$k$ indices of node scores $Z$.
Unlike \textit{pooling by graph transformation}, $G'$ is a sub-graph of $G$ (i.e. $V' \subset V, E' \subset E$).
Because the nodes are explicitly selected, it would be helpful to interpret which local structures are important to increase the predictive power of GNNs.

\subsubsection{Why is CAGPool based on node-selection?}
As referred in Section \ref{subsec:graph_transform} and Section \ref{subsec:node_sel}, pooling by node selection alleviates the computational complexity issue of pooling by graph transformation such as DiffPool.
Since the main goal of our model is to design a low-compleixty model that can effectively and efficiently encode the interaction representation between the pair of input graphs, we follow the architectural design of \textit{pooling by node selection}.
For instance, if we adapt the graph DiffPool-like CAGPool which is described the below section, the complexity is quadratic to the number of nodes for each graph, $\mathcal{O}(|V_A|^2 + |V_B|^2)$.
Also we can identify the important substructure in the original topology that is difficult in \textit{Pooling by graph transformation}.
During the experiments in our main paper, we followed 50\% ratio pooling for fair comparison with other \textit{pooling by node selection} methods.

\subsubsection{Extension of CAGPool to a graph-transform version}
While pooling by node selection can be beneficial in terms of computational complexity and interpretability, it is also true that some nodes are not selected during the pooling process and are therefore discarded. 
On the other hand, pooling by graph transformation includes all nodes in the final aggregated clusters, without loss of information. 
CAGPool can be extended to a graph transformation version by setting the assignment matrix as follows:
\begin{equation}
    S = h(X_A, X_B)
\end{equation}
This means that the clustering is dynamically performed based on the representation of the pair of graphs.

\section{\break Comparison of the running time}
\label{sec:comparison}
To demonstrate the efficiency of our method, we compare the running time of node-level and graph-level interaction modules. 
Each module accepts $X_A$ and $X_B$ as an input pair and outputs $X'_A$ and $X'_B$, which represent the pooled node feature matrix. 
We set the number of nodes from 50 to 200 and repeat the process 10k times to obtain consistent results. The graph-level interaction module produces $X'_A$ and $X'_B$ 31.2 - 64.7 \% faster than the node-level interaction module.

\begin{figure}
\centering
    \includegraphics[width=\columnwidth]{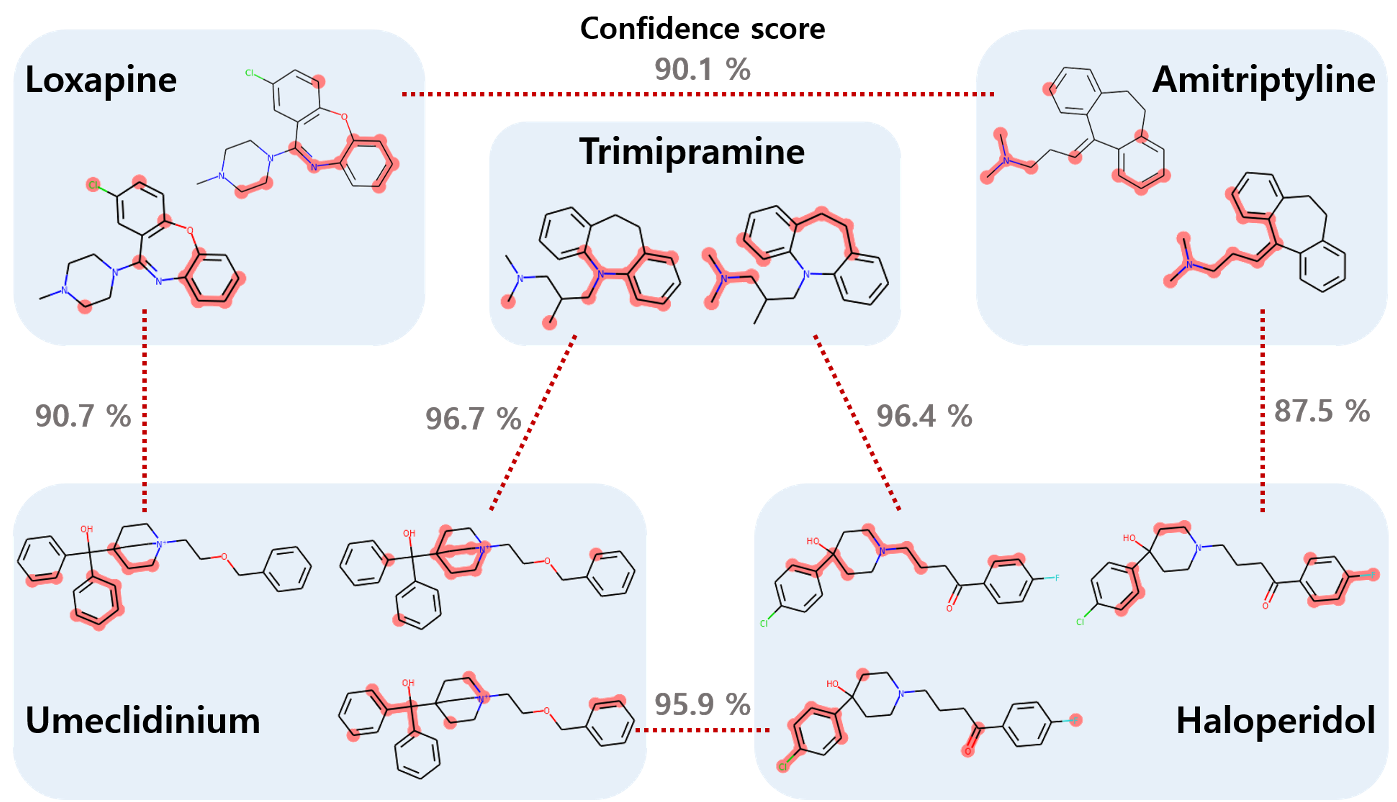}
    \caption{Visualization of the top attention score nodes on true-positive samples from the test set. The highlighted areas represent attention patterns based on the pairs. The percentages indicate our prediction values.}
    \label{fig:fig_visualization}
\end{figure}
\section{\break Visualization of the attention patterns}
\label{sec:comparison}
Figure \ref{fig:fig_visualization} showcases a relationship diagram that includes visualization of attention areas for an example of many-to-many interactions. 
We examined the true-positive cases of our model and highlighted the nodes with the highest attention scores. 
The percentage above the edge between each structure represents the confidence score of our prediction model. 
As shown in the figure, each drug is pooled and projected differently based on the drug it interacts with.


\bibliographystyle{unsrt}
\bibliography{access}

\begin{IEEEbiography}[{\includegraphics[width=1in,height=1.25in,clip,keepaspectratio]{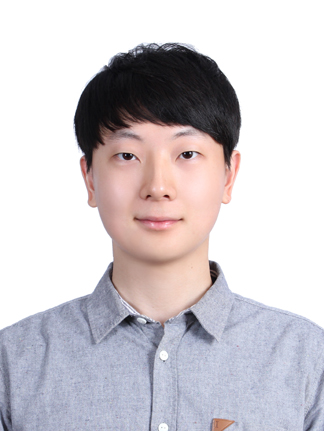}}]{JUNHYUN LEE} received a B.S. degree in Biomedical Engineering from Korea University in 2017. He is currently pursuing a Ph.D. degree in Computer Science at Korea University and supervised by professor Jaewoo Kang.
His research interests include Deep Learning, Geometric Machine Learning, and Biomedical Applications.
\end{IEEEbiography}

\begin{IEEEbiography}[
    {\includegraphics[width=1in,height=1.25in,clip,keepaspectratio]{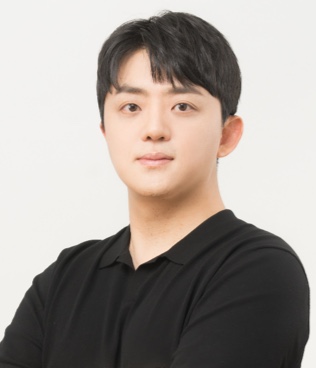}}]
{Bumsoo Kim} received his B.S. degree in Computer Science from Korea University, Seoul, South Korea, in 2016. He completed his Ph.D. in Computer Science also at Korea University in 2022. During his Ph.D., he concurrently worked as a Research Scientist at Kakao Brain, South Korea from 2020 to 2022. After his Ph.D., he is working as a Research Scientist at LG AI Research, South Korea, where his main research interests covers Deep Learning, Scene Understanding, Efficient Transformers, and Large-Scale Vision-Language Pretraining.

\end{IEEEbiography}

\begin{IEEEbiography}[
    {\includegraphics[width=1in,height=1.25in,clip,keepaspectratio]{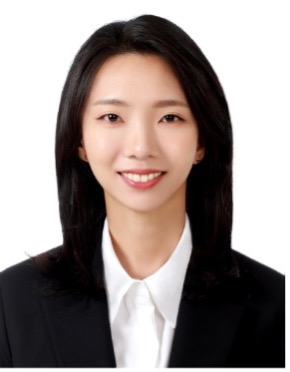}}]
{Minji Jeon} earned her B.S. degree in Computer Science from Korea University, Seoul, South Korea, in 2012. She then received her M.S. degree from the same university in 2014 as part of the Interdisciplinary Graduate Program in Bioinformatics. She completed her Ph.D. in Computer Science, also at Korea University, in 2018. Upon completion of her Ph.D., she held the position of Research Professor at Korea University from 2018 to 2019. She subsequently worked as a Postdoctoral Fellow at the Icahn School of Medicine at Mount Sinai in New York, USA, from 2020 to 2022. As of 2022, she has taken on the role of Assistant Professor within the Department of Medicine at the Korea University College of Medicine in Seoul, South Korea.

\end{IEEEbiography}

\begin{IEEEbiography}[
    {\includegraphics[width=1in,height=1.25in,clip,keepaspectratio]{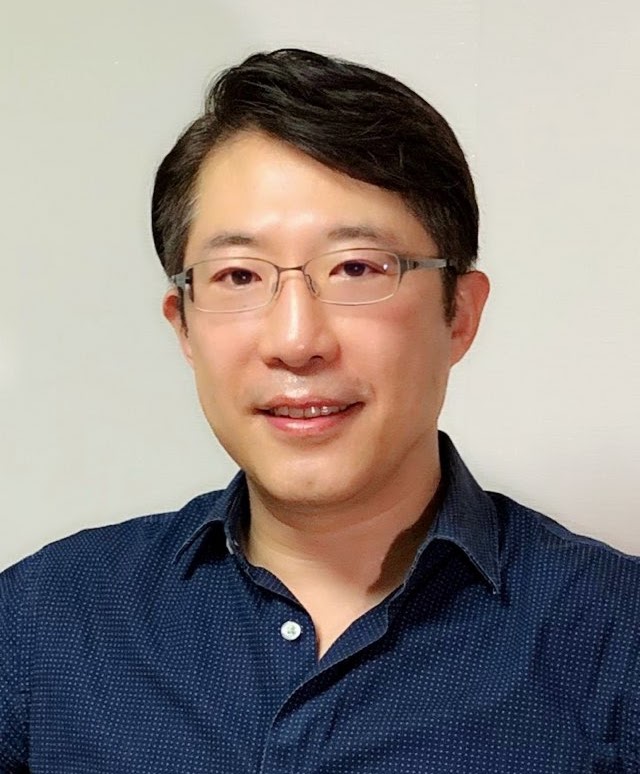}}]
{Jaewoo Kang} received the B.S. degree in computer science from Korea University, Seoul, South Korea, in 1994, the M.S. degree in computer science from the University of Colorado Boulder, CO, USA, in 1996, and the Ph.D. degree in computer science from the University of Wisconsin–Madison, WI, USA, in 2003. From 1996 to 1997, he was a Technical Staff Member at AT\&T Labs Research, Florham Park, NJ, USA. From 1997 to 1998, he was a Technical Staff Member with Savera Systems Inc., Murray Hill, NJ, USA. From 2000 to 2001, he was the CTO and a Co-Founder of WISEngine Inc., Santa Clara, CA, USA, and Seoul. From 2003 to 2006, he was an Assistant Professor with the Department of Computer Science, North Carolina State University, Raleigh, NC, USA. Since 2006, he has been a Professor with the Department of Computer Science, Korea University, where he also works as the Department Head of Bioinformatics for Interdisciplinary Graduate Program.

\end{IEEEbiography}

\EOD

\end{document}